# Pre-Trained Convolutional Neural Network Features for Facial Expression Recognition

Aravind Ravi
Department of Systems Design Engineering
University of Waterloo, Ontario CA
aravind.ravi@uwaterloo.ca

*Abstract*— Facial expression recognition has been an active area in computer vision with application areas including animation, social robots, personalized banking, etc. In this study, we explore the problem of image classification for detecting facial expressions based on features extracted from pre-trained convolutional neural networks trained on ImageNet database. Features are extracted and transferred to a Linear Support Vector Machine for classification. All experiments are performed on two publicly available datasets such as JAFFE and CK+ database. The results show that representations learned from pre-trained networks for a task such as object recognition can be transferred, and used for facial expression recognition. Furthermore, for a small dataset, using features from earlier layers of the VGG19 network provides better classification accuracy. Accuracies of 92.26% and 92.86% were achieved for the CK+ and JAFFE datasets respectively.

## I. Introduction

Facial expressions play a significant role in human communication. They serve as one of the main information channels in interpersonal communications. Facial expression recognition (FER) has been an active area of research in computer vision with application areas including social robots, personalized banking, animation, etc. Traditional computer vision algorithms have focused on extracting hand-crafted features based on methods such as Gabor wavelets, SIFT, local binary patterns, etc. There has been increased attention in a new way of automatically extracting features from data. This method of automatically learning abstract representations or features from the data is called representation learning.

In recent years, there has been an upsurge in the use of convolutional neural networks (CNN) for image classification and object detection. A CNN can be viewed as a framework combining a feature extractor and a classifier. The convolutional layers behave as feature extractors that learn the representations automatically from the input data. Features such as shapes, edges, and color blobs are learned in the earlier layers of the CNN. While the later layers learn features more specific to the original dataset. The learned features are fed into the last fully connected layers to classify the data into one of the classes.

Deep CNN features have been shown to be useful as a preliminary candidate for image classification [7]. Such features provide an alternate approach to existing hand-crafted image features. This technique is popularly known as transfer learning. Transfer learning is defined as the ability of a system to recognize and apply knowledge and skills learned in previous tasks to novel tasks [1]. It aims to extract the knowledge from one or more source tasks and applies the knowledge to a target task. The objective is to learn a good feature representation for the target domain. These feature representations encode the knowledge that is used to transfer across the domains. One approach for transfer learning is to use a CNN, pre-trained on ImageNet [7], as a fixed feature extractor. In this approach, the last fully connected layers are removed, and the rest of the network is used as a fixed feature extractor for the target domain. The features thus obtained can be used to train a linear classifier such as Linear Support Vector Machine (SVM) or softmax classifier for the target dataset. In addition to this, the second approach involves fine-tuning the weights of the CNN by continuing the backpropagation on the target domain. Fine-tuning can be performed on all layers or it is possible to freeze some of the weights of earlier layers and fine-tune deeper layers of the CNN.

There are several factors to be considered while choosing the approach of transfer learning for a given problem. The size of the target dataset and its similarity to the original dataset are two important factors to be considered. Based on these factors, different methods can be adopted [2].

*Target dataset is small and similar to original dataset*
Since the dataset is small and similar to original dataset, fine-tuning the CNN could lead to overfitting. Due to the similarity with the original dataset, higher level features are expected to be relevant to the target dataset. Hence, a linear classifier can be trained on the obtained features.

*Target dataset is large and similar to original dataset*
In this case, since the target dataset is large, the CNN can be fine-tuned as we can expect the target dataset to have more diverse samples.

*Target dataset is small but very different from the original dataset*
Since the target dataset is small, a linear classifier can be trained on the extracted features. As the target dataset is very different from the original dataset, the features extracted from earlier layers can provide generic representations that can be transferred to train a linear classifier.

*Target dataset is large and very different from the original dataset*
Since the target dataset is very large, the CNN can be trained from scratch. But, in practice, the CNN weights are initialized with the weights from the pre-trained network.

Prior studies have been successful in applying transfer learning for tasks such as object recognition [3]. Several studies have reported success when attempting to utilize pre-trained network features for medical imaging tasks [4] [5] [6] [14].

A method for facial expression recognition based on transfer learning techniques is studied in this work. We explore the use of features extracted from the pre-trained VGG19 network [7] trained on ImageNet dataset for detecting facial expressions. Section II introduces the overall system design for detecting the facial expressions. Section III gives a description of the datasets used in this study. In Section IV, the proposed methodology for transferring the features from the pre-trained model and using them to train a linear classifier is discussed. Finally, the results of the experiments performed on the two datasets are discussed in Section V and VI.

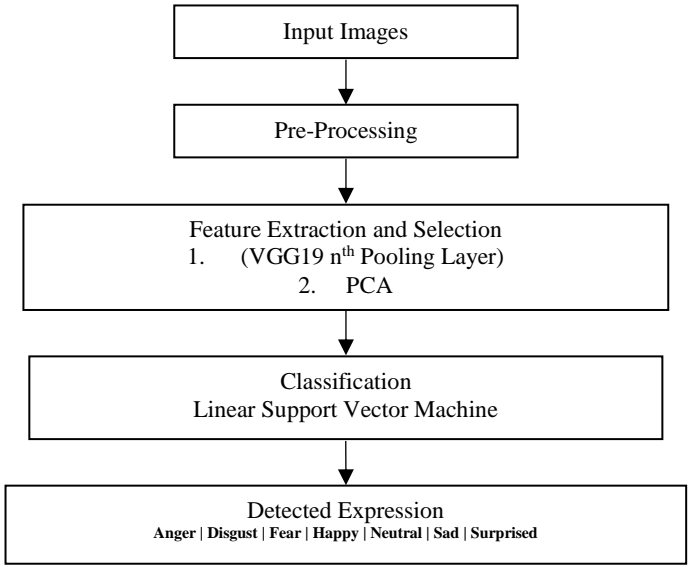

Fig. 1. Overall system design illustrating the steps during training and testing phase.

## II. SYSTEM DESIGN

The proposed system consists of the following steps: pre-processing, feature extraction, feature selection and classification. Further, this consists of two phases of learning: training and testing. During the training phase, the input grayscale images are pre-processed by performing intensity normalization on the pixel values and resizing to 224 by 224 pixels. These images are given as input to the pre-trained VGG19 network. The VGG19 contains five pooling layers as shown in Table 1. The output of each pooling layer and the first fully connected layer are extracted and stored as a set of features for further processing. Therefore, we have six sets of features in total. These features are evaluated further to identify the best features for transfer learning. After extracting the features, principal component analysis (PCA) is performed on each set of feature output for dimensionality reduction and feature selection. Finally, the selected features are used to train a linear SVM classifier to output the class label of the predicted facial expression. The testing phase follows a similar procedure as the training phase, wherein the new test images are first intensity normalized. Following this, the features are extracted from each pooling layer and first fully connected layer. PCA is performed to reduce dimensions and select the features, and the linear SVM classifier predicts the output class for a given test image. The methodology of each of the steps is detailed in Section IV. The overall system is illustrated in Fig. 1.

## III. DATASET DESCRIPTION

The experiments were performed on two publicly available facial expression databases: The Japanese Female Facial Expression (JAFFE) database [8] and the Extended Cohn-Kanade (CK+) database [9]. The JAFFE database consists of 213 images from 10 Japanese female subjects. For each subject, there are around 4 images for each of the seven expressions (including neutral). All images are grayscale images of size 256 by 256 pixels. The database is separated into seven groups based on the seven classes of expressions for this experiment. Some examples of the JAFFE database images are shown in Fig. 2.

TABLE I. VGG19 ARCHITECTURE AND LAYERS CONFIGURATION

| LAYER (TYPE) | OUTPUT SHAPE | PARAMETERS |
|---|---|---|
| INPUT_1 (INPUT LAYER) | (NONE, 224, 224, 3) | 0 |
| BLOCK1_CONV1 (CONV2D) | (NONE, 224, 224, 64) | 1792 |
| BLOCK1_CONV2 (CONV2D) | (NONE, 224, 224, 64) | 36928 |
| **BLOCK1_POOL (MAXPOOLING2D)** | **(NONE, 112, 112, 64)** | **0** |
| BLOCK2_CONV1 (CONV2D) | (NONE, 112, 112, 128) | 73856 |
| BLOCK2_CONV2 (CONV2D) | (NONE, 112, 112, 128) | 147584 |
| **BLOCK2_POOL (MAXPOOLING2D)** | **(NONE, 56, 56, 128)** | **0** |
| BLOCK3_CONV1 (CONV2D) | (NONE, 56, 56, 256) | 295168 |
| BLOCK3_CONV2 (CONV2D) | (NONE, 56, 56, 256) | 590080 |
| BLOCK3_CONV3 (CONV2D) | (NONE, 56, 56, 256) | 590080 |
| BLOCK3_CONV4 (CONV2D) | (NONE, 56, 56, 256) | 590080 |
| **BLOCK3_POOL (MAXPOOLING2D)** | **(NONE, 28, 28, 256)** | **0** |
| BLOCK4_CONV1 (CONV2D) | (NONE, 28, 28, 512) | 1180160 |
| BLOCK4_CONV2 (CONV2D) | (NONE, 28, 28, 512) | 2359808 |
| BLOCK4_CONV3 (CONV2D) | (NONE, 28, 28, 512) | 2359808 |
| BLOCK4_CONV4 (CONV2D) | (NONE, 28, 28, 512) | 2359808 |
| **BLOCK4_POOL (MAXPOOLING2D)** | **(NONE, 14, 14, 512)** | **0** |
| BLOCK5_CONV1 (CONV2D) | (NONE, 14, 14, 512) | 2359808 |
| BLOCK5_CONV2 (CONV2D) | (NONE, 14, 14, 512) | 2359808 |
| BLOCK5_CONV3 (CONV2D) | (NONE, 14, 14, 512) | 2359808 |
| BLOCK5_CONV4 (CONV2D) | (NONE, 14, 14, 512) | 2359808 |
| **BLOCK5_POOL (MAXPOOLING2D)** | **(NONE, 7, 7, 512)** | **0** |
| FLATTEN (FLATTEN) | (NONE, 25088) | 0 |
| **FC1 (DENSE)** | **(NONE, 4096)** | **102764544** |
| FC2 (DENSE) | (NONE, 4096) | 16781312 |
| PREDICTIONS (DENSE) | (NONE, 1000) | 4097000 |

Another database used to evaluate the proposed system is the Extended Cohn-Kanade (CK+) database. This comprises of 100 university students with age between 18 and 30 years old. The subjects formed a diverse group of individuals comprising of both male and female subjects of ethnicity belonging to one of Asian, African-American, American or South American. The students were instructed to perform a series of expressions and the images were captured by a camera placed directly in front of the subjects. All images are grayscale of size 640 by 480 pixels. A subset of the CK+ database with 10 subjects, 5 male and 5 female for each of the seven expressions resulting in a total of 210 images were selected for this study. Some examples of the CK+ database images are shown in Fig. 2. In both datasets, the seven expressions were: anger, happy, sad, disgust, fear, surprised and neutral.

## IV. METHODOLOGY

All experiments were performed on the pre-trained VGG19 networks provided in the Keras package for Python [10]. This framework is used in evaluating the performance of the CNN as a feature extractor, and for transferring the features for facial expression recognition.

### A. Feature Extraction

Table 1 summarizes the layer configurations of the VGG19 network. The features extracted from all the pooling layers (block1_pool, block2_pool, block3_pool, block4_pool, and block5_pool) and the first fully connected layer (fc1) of the VGG19 are evaluated using the implementation provided for specific architectures in Keras. Each layer's output is vectorized and used as the feature vector.

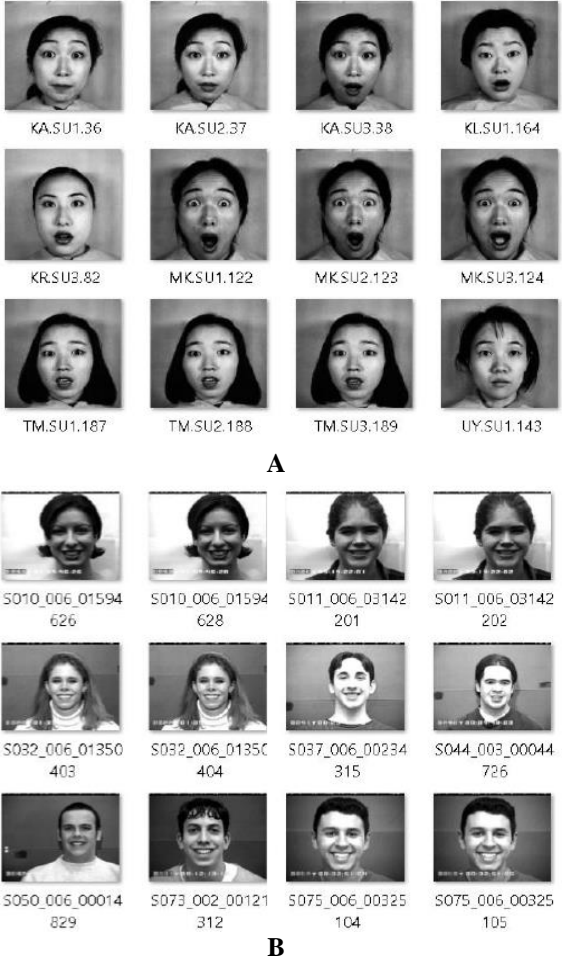

Fig. 2A. Examples of the images in the JAFFE database. Fig. 2B. Examples of the images in the CK+ database.

Table 2 summarizes the feature vector sizes of each pooling layer's output. It is not practical to use the feature vectors as it is to train a classifier as the size of the feature vector is large. Hence, reduction in dimensionality is required. To reduce the dimensionality of the feature vectors, PCA is applied to each pooling layer and fully connected layer's output. To choose the number of PCA components ($N_{PCA}$) different values such as $N_{PCA} = \{50, 100, 150, 200\}$ were evaluated.

*B. Feature Selection*

The extracted features from the previous step are transformed to a lower dimensional vector by applying PCA transformation based on the chosen $N_{PCA}$ value. Feature selection for the final classifier is based on two parameters:

1. $N^{th}$ VGG19 layer for feature extraction
2. $N_{PCA}$ value

The parameter selection is based on a two-step method. The first step is to select those parameters that provide the two highest accuracies on the training set based on a Jack-Knife validation ($A_{JK}$). Next, select the final parameters based on the parameters that have the least difference between the $A_{JK}$ and corresponding test accuracy ($A_T$). Using this method, the $N_{PCA}$ and the $N^{th}$ layer of VGG19 for feature extraction are selected.

*C. Classification*

As discussed earlier, for transfer learning, since the target dataset is different from the source dataset and also smaller in size, a linear classifier is trained on the features extracted from the pre-trained CNN. A Support Vector Machine with a linear kernel is used as the classifier. The python package sci-kit-learn [11] based on the LIBLINEAR [12] implementation was used to train the SVM and NumPy [13] was used to process and store the data during the experiments.

*D. Validation*

The training data for both datasets were split into 80% for training the classifier and 20% was used for testing. These results are validated based on a 10-fold cross-validation, and due to the small size of the datasets, a Jack-Knife validation or leave-one-out validation is also performed. The validation results are summarized in Fig. 3. Based on the validation results, the highest accuracies for JAFFE and CK+ datasets were achieved for $N_{PCA} = 200$ and $N_{PCA} = 100$ respectively.

## V. RESULTS

Table 3 summarizes the training and test accuracies for the chosen number of $N_{PCA}$. Among the different methods of feature extraction, features from Block4 pool layer of VGG19 provides the highest accuracy for both CK+ and JAFFE dataset. Applying the proposed feature selection methods, an $A_{JK}$ of 92.77% and $A_T$ of 92.86% was achieved for the JAFFE dataset. An $A_{JK}$ of 92.26% and $A_T$ of 92.86% was achieved on the subset of the CK+ dataset. It can be observed from the $A_{JK}$ values wherein earlier layers of the CNN provide a better performance than later layers.

TABLE II. FEATURE VECTOR SIZE PER LAYER

| LAYER | FEATURE VECTOR SIZE |
|---|---|
| BLOCK1_POOL | 802816 |
| BLOCK2_POOL | 401408 |
| BLOCK3_POOL | 200704 |
| BLOCK4_POOL | 100352 |
| BLOCK5_POOL | 25088 |
| FC1 | 4096 |

TABLE III. TRAINING AND TEST ACCURACIES FOR SELECTED FEATURES OF CK+ AND JAFFE DATASETS

| | CK+ DATASET | | | JAFFE DATASET | | |
|---|---|---|---|---|---|---|
| FEATURES | PCA - 100 | | | PCA – 200 | | |
| VGG19 LAYER (POOL) | TRAINING (80%) | | TEST (20%) | TRAINING (80%) | | TEST (20%) |
| | 10-FOLD | $A_{JK}$ | | 10-FOLD | $A_{JK}$ | |
| BLOCK1 | 87.90 | 88.69 | 85.71 | 77.27 | 79.52 | 88.10 |
| BLOCK2 | 90.27 | 90.48 | 85.71 | 81.66 | 83.73 | 88.10 |
| BLOCK3 | 91.51 | 93.45 | 90.48 | 90.47 | 89.76 | 92.86 |
| **BLOCK4** | **94.93** | **92.26** | **92.86** | **92.70** | **92.77** | **92.86** |
| BLOCK5 | 88.83 | 90.48 | 90.48 | 81.35 | 82.53 | 85.71 |
| (DENSE) FC1 | 91.76 | 89.88 | 92.86 | 81.01 | 76.51 | 88.10 |

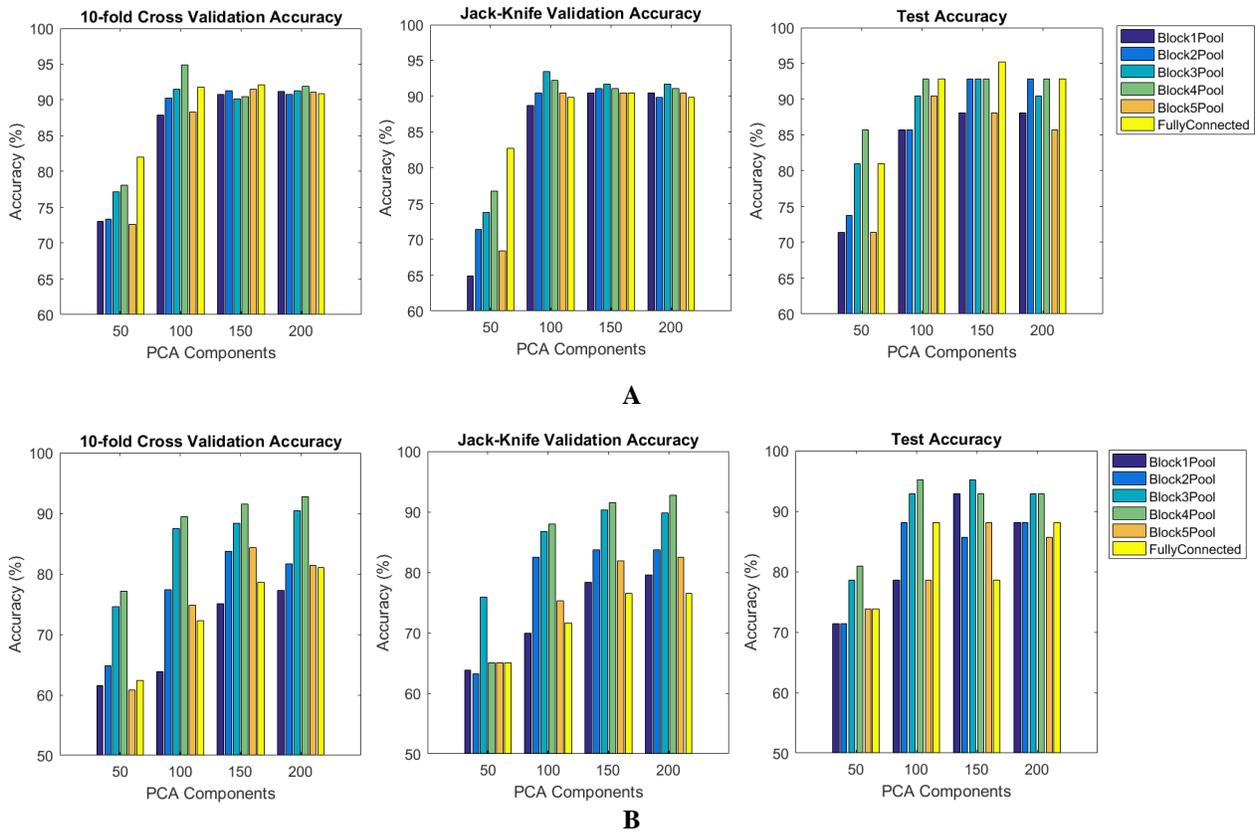

Fig 3. Training and Test accuracies for features extracted from different layers of VGG19 for selected features based on PCA components on A. CK+ Dataset (Top), B. JAFFE Dataset (Bottom)

## VI. CONCLUSION

This paper presents an approach for using pre-trained convolutional neural network features for facial expression recognition. One of the popular approaches using transfer learning is studied in this work wherein the learned representations of a pre-trained CNN trained for one particular task can be used for a novel task. A pre-trained VGG19 network trained on ImageNet database is evaluated for facial expression recognition. The CNN is used as a feature extractor and an SVM classifier with a linear kernel is trained on these features to predict the facial expression. Experiments were performed on two publicly available datasets such as JAFFE and CK+. The results suggest that representations learned from pre-trained networks trained for a particular task such as object detection can be transferred, and used for a different task such as facial expression recognition. Furthermore, for a small dataset, using features from earlier layers of the network provide better accuracy.